\documentclass{article}

\usepackage{spconf}
\usepackage{amsmath}
\usepackage{graphicx}

\usepackage{amssymb}
\usepackage{algorithm}
\usepackage{algorithmic}
\usepackage{amsfonts}
\usepackage{caption}
\usepackage{color}
\usepackage{epsfig}
\usepackage{multirow}
\usepackage{tablefootnote}
\usepackage{times}

\name{Jie Cai$^{\star}$, Zibo Meng$^{\star}$, Ahmed Shehab Khan$^{\dagger}$,  James O'Reilly$^{\dagger}$, Zhiyuan Li$^{\dagger}$, Shizhong Han$^{\ast}$, and Yan Tong$^{\dagger}$}

\address{$^{\star}$ InnoPeak Technology, $^{\dagger}$ University of South Carolina,  $^{\ast}$ Qualcomm AI Research}

\begin{document}

\title{Identity-Free Facial Expression Recognition using Conditional Generative Adversarial Network}

\maketitle

\begin{abstract}

A novel Identity-Free conditional Generative Adversarial Network (IF-GAN) was proposed for Facial Expression Recognition (FER) to explicitly reduce high inter-subject variations caused by identity-related facial attributes, e.g., age, race, and gender. As part of an end-to-end system, a cGAN was designed to transform a given input facial expression image to an ``average'' identity face with the same expression as the input. Then, identity-free FER is possible since the generated images have the same synthetic ``average'' identity and differ only in their displayed expressions. Experiments on four facial expression datasets, one with spontaneous expressions, show that IF-GAN outperforms the baseline CNN and achieves state-of-the-art performance for FER.

\end{abstract}

\begin{keywords}
Facial Expression Recognition, Facial Expression Generation, Convolutional Neural Network, Generative Adversarial Network
\end{keywords}

\section{INTRODUCTION}
\begin{figure}[th]
   \centering
   \includegraphics[width=0.5\textwidth]{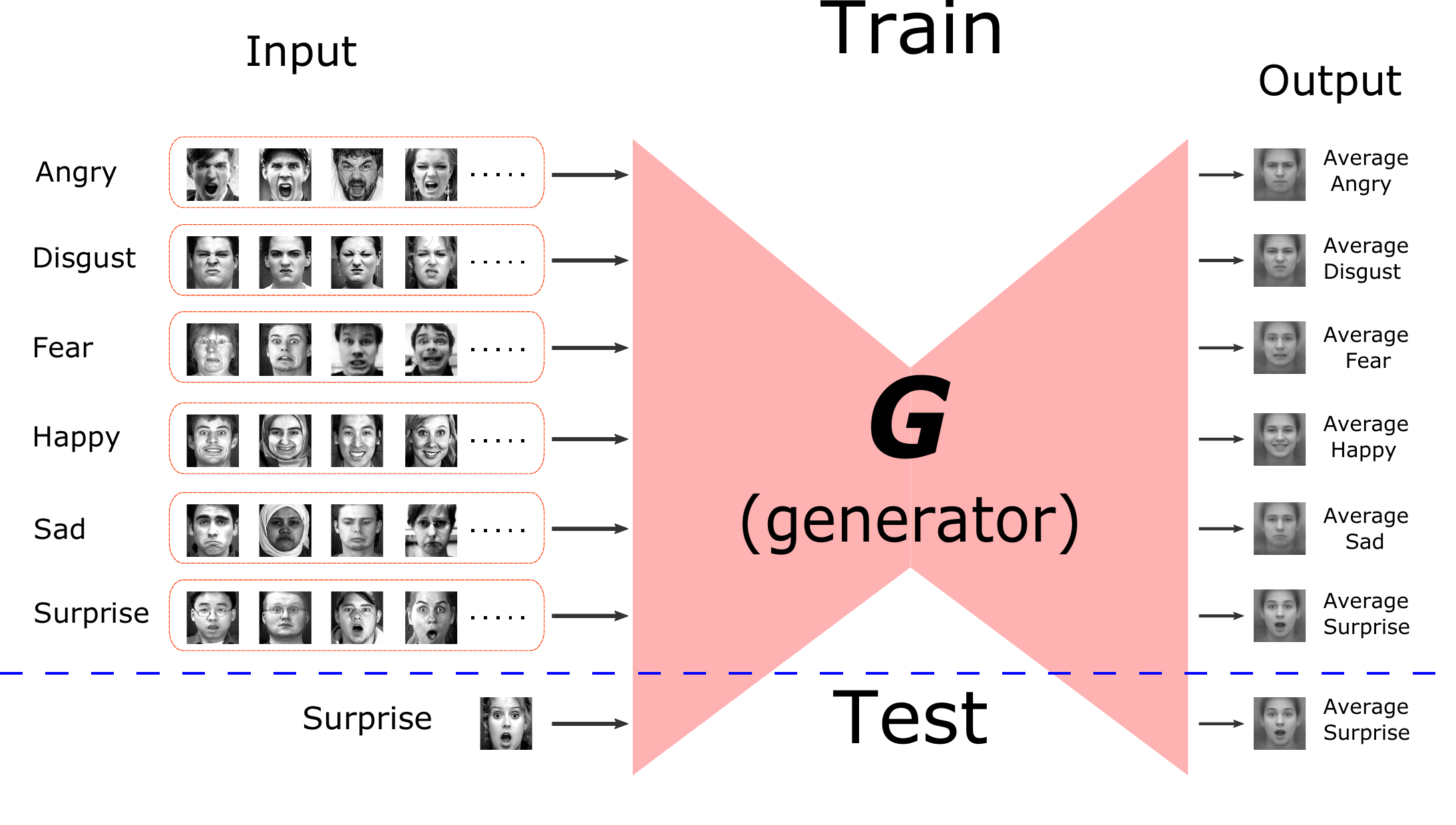}
   \caption{\small{IF-GAN identity-free expressive face generation.}}
   \label{fig:introduction}
\end{figure}

FER systems play an important role in human-computer interaction, e.g., animation and interactive games.
CNNs have recently achieved good results on FER but the high inter-subject facial attribute variations were not explicitly considered and, as discussed in~\cite{yang2018identity}, the learned features may capture more identity-related information which is unrelated to the FER task.
This motivates reducing identity-related variations by explicitly removing the identity information from the images.
If $f(\mathbf{I})$ is a CNN extracted facial representation of an input image $\mathbf{I}$ and is the result of a nonlinear function $g(f_{_{id}}(\mathbf{I}), f_{_{exp}}(\mathbf{I})) \to f(\mathbf{I})$, where $f_{_{id}}(\mathbf{I})$ are identity-related features and $f_{_{exp}}(\mathbf{I})$ are expression-related features, then one may focus on $f_{_{exp}}(\mathbf{I})$ and minimize $f_{_{id}}(\mathbf{I})$ when finding $f(\mathbf{I})$.

Fig.~\ref{fig:introduction} illustrates the proposed IF-GAN removing identity features by transferring the expression features from an input image to a synthetic ``average'' face calculated from all subjects in a training dataset. The generated images have the same synthetic ``average'' identity, differing only in their displayed expressions, and thus can be used for identity-free expression classification.

In summary, our major contributions are:
\begin{enumerate}
\item[-] Identity-free FER using IF-GAN by transferring an arbitrary subject's expression to an ``average'' identity;
\item[-] Developing an end-to-end system to perform expression synthesis and recognition simultaneously.
\end{enumerate}

IF-GAN surpasses baseline CNN and state-of-the-art FER methods in extensive experiments on four facial expression datasets, including the spontaneous facial expression dataset RAF-DB.

\section{RELATED WORK}
Recent surveys~\cite{Martinez2017Automatic,li2018deep} detail the research into FER over past decades.
Deep CNNs have recently achieved good FER results~\cite{meng2017identity,cai2018island,zhao2016peak,li2018reliable,lopes2017facial,a2018covariance,ding2017facenet2expnet,zeng2018facial,
yang2018facial,yang2018identity,chen2020label,fu2019ferlrtc,li2019Pooling,lipatch,li2019occlusion,perveen2020facial,
wang2020region,wang2020suppressing}, but they may also learn identity-related features that are irrelevant to expression and suffer from high intra-class variations and inter-class similarities, leading to a drop in FER performance on unseen subjects.
Wen et al.~\cite{wen2016discriminative} introduced a center loss for face recognition to reduce intra-class variations, without explicitly considering inter-class similarity. Extensions of center loss~\cite{cai2018island,li2018reliable} were proposed to reduce the intra-class variations while increasing the inter-class differences. These approaches, however, are trained on distinct subjects without disentangling identity-related and expression-related features.

\textbf{Identity-free facial expression recognition:}
A few recent approaches attempted to explicitly improve person-independent FER. An Identity-Aware CNN (IACNN)~\cite{meng2017identity} was proposed to alleviate identity-related variations using expression-sensitive and identity-sensitive contrastive losses. Drastic data expansion with contrastive loss occurs when constructing image pairs for training.
Identity-Adaptive Generation (IA-gen)~\cite{yang2018identity} was developed using six cGANs to generate six expressions from any input facial image. Then, FER is performed by comparing the six output images to the input.
In De-expression Residue Learning (DeRL)~\cite{yang2018facial}, a cGAN synthesized a neutral, same-identity face image from a given expressive image. Person-independent expression information was extracted from the intermediate layers of the generative model.
A cGAN was utilized in~\cite{chen2018vgan} to transform an input face image $\mathbf{I}$ to an expression-preserving face image $\tilde{\mathbf{I}}$ of a different identity. The generator produced an identity-invariant representation $f(\mathbf{I})$, which was employed for FER.
Unlike these cGAN-based models~\cite{yang2018identity,yang2018facial,chen2018vgan}, the proposed IF-GAN is an end-to-end network that directly removes the identity information and generates a face image for identity-free FER.

\section{METHODOLOGY}
\begin{figure}[th!]
   \centering
   \includegraphics[width=0.5\textwidth]{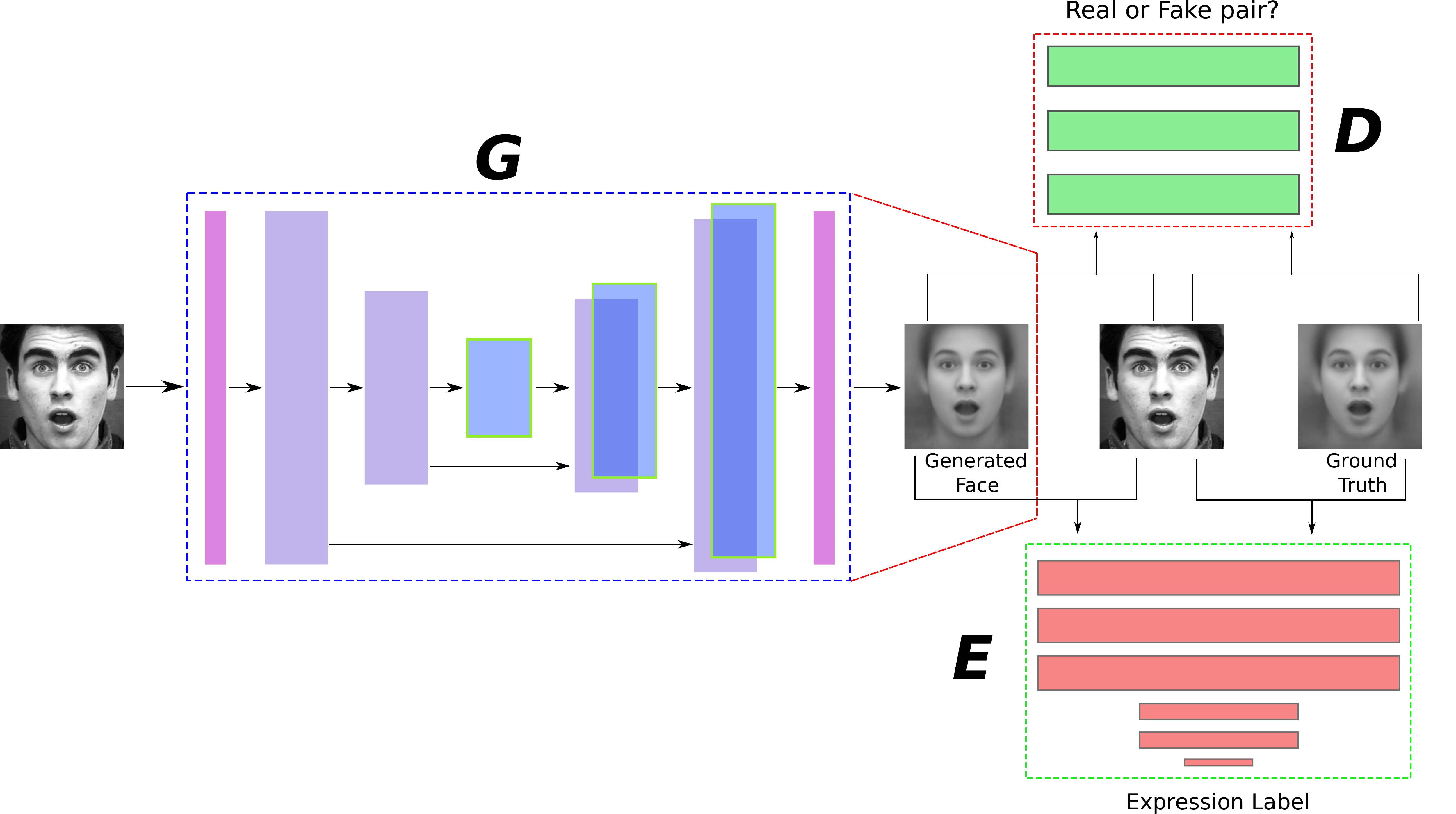}
   \caption{\small{The IF-GAN architecture consists of: (1) a ``U-Net''~\cite{ronneberger2015u} generator ($\mathbf{G}$), (2) a ``PatchGAN''~\cite{isola2017image} discriminator ($\mathbf{D}$) that only penalizes structures at the scale of patches, and (3) a ResNet-101~\cite{he2016deep} expression classifier ($\mathbf{E}$), pre-trained on ImageNet. In contrast to DeRL~\cite{yang2018facial} and IA-gen~\cite{yang2018identity}, $\mathbf{G}$, $\mathbf{D}$, and $\mathbf{E}$ are jointly optimized during the IF-GAN training. Note that only $\mathbf{G}$ and $\mathbf{E}$ are employed during testing.}}
   \label{fig:model}
\end{figure}

\begin{figure}[th!]
   \centering
   \includegraphics[width=0.5\textwidth]{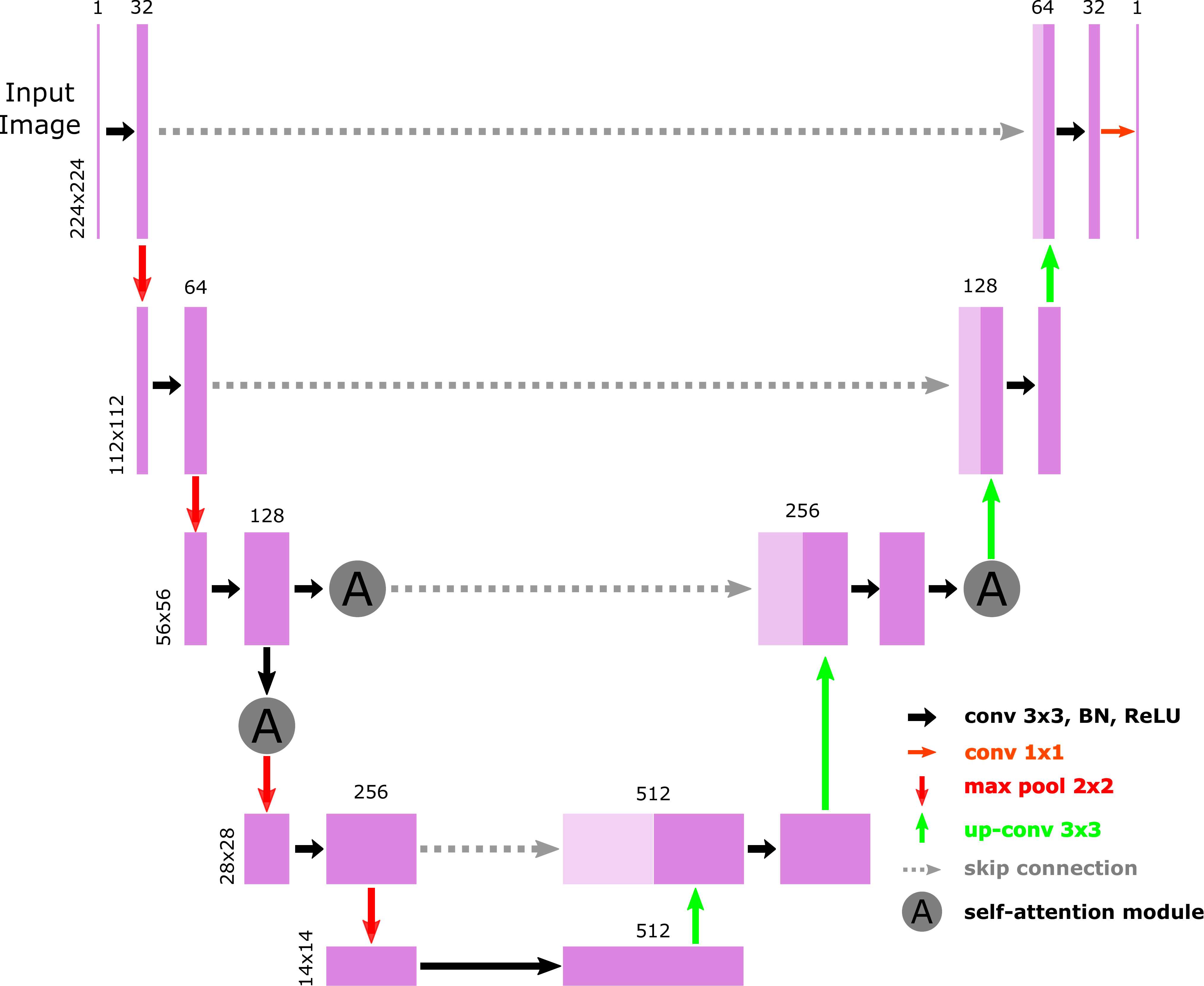}
   \caption{\small{U-Net-based generator with self-attention modules. Purple boxes correspond to multi-channel convolution feature maps; lilac boxes are feature maps copied from the encoding path, where the number of channels is above each box and the spatial size of feature maps are on the box’s lower left corner. Gray dashed arrows denote skip connection operations. Note the three self-attention modules: one on a skip connection and another two on the encoding and decoding paths. Best viewed in color.}}
   \label{fig:U-Net}
\end{figure}

\subsection{An Overview of the Proposed IF-GAN}

The IF-GAN transfers expression information from an input image to a synthetic ``average'' identity and then generates an ``average'' expressive image for identify-free expression classification.

Let $\mathbf{I}_{_{SE}}$ be a face image of any real subject, with any expression, and let $\mathbf{I}_{_{AE}}$ be the ``\textbf{A}verage \textbf{E}xpression'' image calculated from all subjects with the same expression. The generator ($\mathbf{G}$) learns to transfer expression information $f_{exp}(\mathbf{I}_{_{SE}})$ from input $\mathbf{I}_{_{SE}}$ to an ``average'' subject and generates an image $\tilde{\mathbf{I}}_{_{AE}}$ with the same expression as $\mathbf{I}_{_{SE}}$. In contrast to regular GANs and cGANs, an expression classifier ($\mathbf{E}$) is employed in IF-GAN to ensure $\tilde{\mathbf{I}}_{_{AE}}$ has the same expression as the input image $\mathbf{I}_{_{SE}}$.

As shown in Fig.~\ref{fig:model}, the IF-GAN has a generator ($\mathbf{G}$), a discriminator ($\mathbf{D}$), and an expression classifier ($\mathbf{E}$). $\mathbf{G}$ takes an input subject expressive image $\mathbf{I}_{_{SE}}$ and generates an expressive ``average'' subject face image $\tilde{\mathbf{I}}_{_{AE}}$. Then, the fake tuple $\{ \mathbf{I}_{_{SE}}, \tilde{\mathbf{I}}_{_{AE}}, 0\}$ and the real tuple $\{ \mathbf{I}_{_{SE}}, \mathbf{I}_{_{AE}}, 1\}$ are fed into the discriminator for fake/real classification. Finally, $\{\mathbf{I}_{_{SE}}, \mathbf{I}_{_{AE}}, l_e\}$ and $\{\mathbf{I}_{_{SE}}, \tilde{\mathbf{I}}_{_{AE}}, l_e\}$, with the expression label $l_e$, are utilized to fine-tune the expression classifier.

\begin{figure}[th!]
   \centering
   \includegraphics[width=0.5\textwidth]{./attention.pdf}
   \caption{\small{Illustration of the self-attention module. Input feature maps are shown with their tensor shapes, e.g., $C_{1} \times W \times H$ for $C_{1}$ channels of $W \times H$ feature maps. Green/red boxes are $1\times1$ convolution channel reductions resulting in $C_{2}$ channels. Magenta boxes are $1\times1$ convolutions with $C_{1}$ channel results. Softmax, performed on each row, normalizes the attention coefficients. Trainable parameter ($\gamma$) scales self-attention feature maps. Best viewed in color.}}
   \label{fig:attention}
\end{figure}

Furthermore, self-attention modules are added on top of our generator (Fig.~\ref{fig:U-Net}) to efficiently capture spatially distant relationships and generate better expressive images. Fig.~\ref{fig:attention} shows a self-attention module computing a response at a position as a weighted sum of the features at all positions in the input feature maps.

\subsection{Loss Functions of the IF-GAN}

The loss function of the IF-GAN is given below:
\begin{footnotesize}
\begin{equation} \label{eq:joint_IF-GAN}
 \begin{aligned}
\mathcal{L} & =\lambda_{1} \cdot \mathcal{L}_{cGAN}(\mathbf{G},\mathbf{D}) + \lambda_{2} \cdot \mathcal{L}_{L1}(\mathbf{G})
\\ &
+ \lambda_{3} \cdot \mathcal{L}_{softmax}(\mathbf{E}) + \lambda_{4} \cdot \mathcal{L}_{IL}(\mathbf{E})
 \end{aligned}
\end{equation}
\end{footnotesize}
where the hyperparameters $\lambda_{1}$, $\lambda_{2}$, $\lambda_{3}$, and $\lambda_{4}$~\footnote{In our experiments, we set $\lambda_{1}=1$, $\lambda_{2}=20$, $\lambda_{3}=10$, and $\lambda_{4}=1$  empirically.} are used to balance the four terms.

The cGAN loss function $\mathcal{L}_{cGAN}(\mathbf{G},\mathbf{D})$ is defined as:
\begin{footnotesize}
\begin{equation} \label{eq:cGAN1}
 \begin{aligned}
\mathcal{L}_{cGAN}(\mathbf{G},\mathbf{D}) & =  {\mathbb{E}} [log(\mathbf{D}(\{ \mathbf{I}_{_{SE}}, \mathbf{I}_{_{AE}}\}))]
\\ &
+ {\mathbb{E}} [log(1-\mathbf{D} ( \{ \mathbf{I}_{_{SE}}, \mathbf{G}(\mathbf{I}_{_{SE}})\})  )]
 \end{aligned}
\end{equation}
\end{footnotesize}
where $\mathbf{G}(\mathbf{I}_{_{SE}})$ is the generated image $\tilde{\mathbf{I}}_{_{AE}}$, $\{\mathbf{I}_{_{SE}}, \mathbf{G}(\mathbf{I}_{_{SE}})\}$ is a fake tuple, and $\{\mathbf{I}_{_{SE}}, \mathbf{I}_{_{AE}}\}$ is a real tuple.

Competing against the discriminator $\mathbf{D}$, $\mathbf{G}(\cdot)$ learns the true data distribution using $L1$ distance:
\begin{footnotesize}
\begin{equation} \label{eq:L1}
\mathcal{L}_{L1}(\mathbf{G})={\mathbb{E}} [\lVert \mathbf{I}_{_{AE}}-\textbf{G}( \mathbf{I}_{_{SE}}) \rVert_{1}]
\end{equation}
\end{footnotesize}

$\mathbf{E}$ is jointly trained with $\mathbf{D}$ and $\mathbf{G}$, using softmax loss and expression label $l_e$ to define expression loss:
\begin{footnotesize}
\begin{equation} \label{eq:exp}
 \begin{aligned}
\mathcal{L}_{softmax}(\mathbf{E}) & = {\mathbb{E}} [log \ p(l_e|\{\mathbf{I}_{_{SE}}, \mathbf{I}_{_{AE}}\})]
\\ &
+ {\mathbb{E}} [log \ p(l_e|\{\mathbf{I}_{_{SE}}, \textbf{G}(\mathbf{I}_{_{SE}})\})]
 \end{aligned}
\end{equation}
\end{footnotesize}

Island loss $\mathcal{L}_{IL}(\mathbf{E})$~\cite{cai2018island}, which simultaneously reduces the intra-class variations and increases the inter-class differences, is employed jointly with the softmax loss to train expression classifier $\mathbf{E}$:

\begin{footnotesize}
\begin{equation} \label{eq:island_loss_forward}
\mathcal{L}_{IL}(\mathbf{E}) =  \sum\limits_{i=1}^{M} \| \textbf{x} _{i} - \textbf{c} _{y_{i}} \|^{2}_{2} + \alpha  \sum\limits_{j=1}^{N} \sum\limits_{ \substack{k=1 \\ k \neq j }}^{N} \left(\frac{\textbf{c}_{k} \cdot \textbf{c}_{j}}{ \|\textbf{c}_{k}\|_{_2} \|\textbf{c}_{j}\|_{_2} } +1 \right)
\end{equation}
\end{footnotesize}
where $M$ is the number of images; $N$ is the number of expression classes; $\textbf{c}_{k}$ and $\textbf{c}_{j}$ denote the $k^{th}$ and $j^{th}$ expression centers; and $(\cdot)$ denotes dot product; $\alpha$ set to 0.03 empirically balances the two terms. Minimizing the island loss pushes same-class samples closer while separating them from other samples.

\section{EXPERIMENTS}
\subsection{Experimental Datasets}

\textbf{BU-3DFE}~\cite{yin20063d}: Following~\cite{yang2018identity,yang2018facial}, we employed 1,200 images of six basic expressions (anger, disgust, fear, happiness, sadness, and surprise) with high intensity, i.e., the last two levels.

\textbf{CK+}~\cite{Kanade2000,Lucey2010}, with 118 subjects, has 327 videos labeled with one of seven expressions, i.e., contempt and six basic expressions. As in~\cite{cai2018island}, the last three frames from 309 videos labeled as one of six basic expressions formed a 927-image experimental dataset.

\textbf{MMI}~\cite{pantic2005web} has 208 sequences of 31 subjects displaying six basic expressions in frontal-view. Like~\cite{meng2017identity}, we collected three peak frames in the middle, for a total of 624 images.

\textbf{RAF-DB}~\cite{li2018reliable} is a large spontaneous FER dataset of thousands of subjects with various head poses, lighting conditions, and occlusions. We used the RAF-DB single-label subset (12,271 training images and 3,068 testing images) labeled with one of neutral or six basic expressions.

For the posed facial expression datasets, we used a 10-fold cross-validation strategy, i.e., each dataset was split into 10 subsets, mutually exclusive by subjects. For each run, there were 8 training, one validation, and one testing subsets. Reported results are the average of 10 runs. RAF-DB experiments used the standard training and testing sets.

\subsection{Preprocessing}
We employed the given cropped RAF-DB face images. BU-3DFE, CK+, and MMI face regions were aligned based on the eye centers and nose tip, detected by MTCNN~\cite{zhang2016joint} and then scaled to $256\times 256$ with random horizontal flipping and random -5$^\circ$ and 5$^\circ$ rotations before randomly cropping to $224\times 224$ patches during training. The center-cropped patches are used in testing.

\begin{table}[th!]
  \begin{center}
  	\caption{\small{Performance comparison on three posed datasets.}}
	\scalebox{1}{
    \begin{tabular}{c|c|c|c}
    \hline
    Method                                         &    BU-3DFE         &     CK+          &    MMI   \\
    \hline
    \hline
    Center Loss~\cite{wen2016discriminative}       &    --	            &   92.25          &  73.40  \\
    Island Loss~\cite{cai2018island}               &    --	            &   94.39          & \textbf{74.68}  \\
    IACNN~\cite{meng2017identity}                  &    --	            &   95.37          &  71.55   \\
    DLP-CNN~\cite{li2018reliable}                  &    --	            &   95.78          &  --   \\
    FN2EN~\cite{ding2017facenet2expnet}            &    --	            &   96.80          &  --   \\
    IA-gen~\cite{yang2018identity}                 &    76.83	        &   96.57          &  --   \\
    Lopes et al.~\cite{lopes2017facial}            &    72.89	        &   96.76          &  --  \\
    APM~\cite{li2019Pooling}                       &     --             &    --            &  74.04   \\
    PPDN~\cite{zhao2016peak}     		           &    --	            &  \textbf{97.3}   &  --   \\
    FERLrTc~\cite{fu2019ferlrtc}                   &    82.89           &  --              &  --   \\
    uGMM-MIK~\cite{perveen2020facial}              &    --	            &  --              &  73.2  \\
    DeRL~\cite{yang2018facial}                     &    \textbf{84.17}	&  \textbf{97.3}   &  73.23  \\
    \hline
    \hline
    ResNet-101            &    82.5               &  94.82                   &     71.15              \\
    IF-GAN w/o attention  &    84.25	          &  95.79                   &     73.56              \\
    \textbf{IF-GAN}       &    \textbf{85.25}	   &  \textbf{97.52}         &  \textbf{75.48}        \\
    \hline
    \end{tabular}}
    \label{tab:results_three}
  \end{center}
\end{table}

\subsection{CNN Implementation Details}

ResNet-101~\cite{he2016deep}, pre-trained on ImageNet~\cite{russakovsky2015imagenet}, was our baseline model~\footnote{To make a fair comparison, the island loss is also employed.} and also the expression classifier $\mathbf{E}$ in the IF-GAN. For evaluation, the baseline and IF-GAN were fine-tuned on each dataset's training set, with the other three datasets used as additional training data. For each 10-fold evaluation run for the posed FER datasets, the six ``average'' identity expressive images were obtained by averaging all same-label expressive images in its training set.

With significant pose variations in RAF-DB, the ``average'' identity expressive/neutral images were obtained by averaging all images with the same expression label in the three posed datasets containing only frontal view faces. IF-GAN thus achieves both pose-invariant and identity-free FER.

The baseline and IF-GAN were implemented in PyTorch~\cite{paszke2017automatic} and optimized with Adam~\cite{kingma2014adam} with mini-batch size 50, $\beta1=0.9$, $\beta2=0.99$, and a weight decay parameter of 1e-4 over 100 epochs. Learning rate $\mu$, initially 1e-4, was reduced by a 0.9 factor every 10 epochs.

\subsection{Experimental Results}

\begin{table}[th!]	
  \begin{center}
    \caption{\small{Performance  comparison on the RAF-DB.}}
    \label{tab:results_raf1}
    \begin{tabular}{c|c}
    \hline
    Method                                                &  Accuracy              \\\hline \hline
    Island Loss~\cite{cai2018island}                           &   83.64    \\
    Center Loss~\cite{li2018reliable}                          &   83.68    \\
    DLP-CNN~\cite{li2018reliable}                             &   84.13    \\
    gACNN~\cite{li2019occlusion}                             &   85.07    \\
    APM~\cite{li2019Pooling}                                    &   85.17    \\
    LDL-ALSG~\cite{chen2020label}								   &   85.53    \\
    IPA2LT(LTNet)~\cite{zeng2018facial}                       &   86.77    \\
    RAN~\cite{wang2020region}                                      &   86.90    \\
    Covariance Pooling~\cite{a2018covariance}           &   87.00    \\
    SCN	~\cite{wang2020suppressing} 				        &   \textbf{88.14}    \\
    \hline
    \hline
    ResNet-101                              &     86.31	             \\
    IF-GAN w/o attention                    &     87.32      	     \\
    \textbf{IF-GAN}                         &    \textbf{88.33}	     \\
    \hline
    \end{tabular}
    \label{tab:results_raf}
  \end{center}
\end{table}

As shown in Table~\ref{tab:results_three} and~\ref{tab:results_raf}, IF-GAN with self-attention module outperforms the baseline ResNet-101 and IF-GAN without self-attention module, and achieves state-of-the-art results on all four datasets versus other methods in comparison.  On CK+, DeRL~\cite{yang2018facial} and PPDN~\cite{zhao2016peak} reach state-of-the-art performance, but DeRL~\cite{yang2018facial} is not end-to-end trainable and thus incurs high computational cost while PPDN~\cite{zhao2016peak} uses reference neutral images, which are not available for most real-world applications.

\subsection{Expression Transfer Analysis}

\begin{figure}[th]
   \centering
   \includegraphics[width=0.5\textwidth]{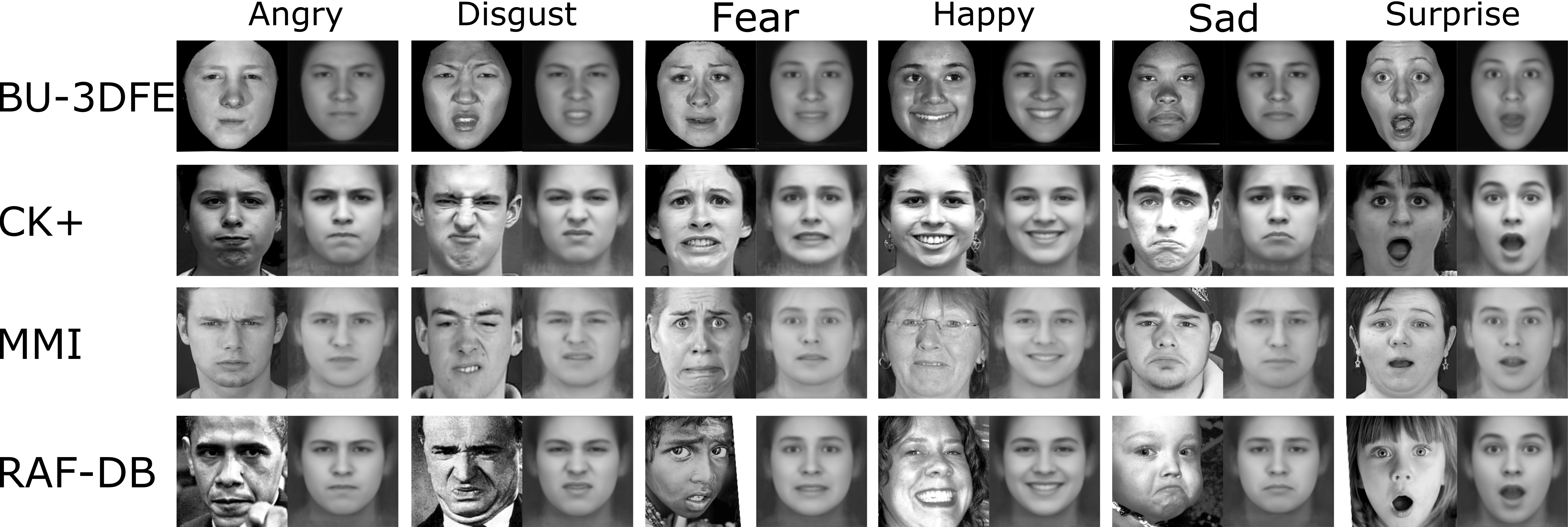}
   \caption{\small{Examples of expression transfer results on four datasets.}}
   \label{fig:results}
\end{figure}

Fig.~\ref{fig:results} demonstrates ``average'' identity expressive image generation from input images. Generated face images completely remove identity information of the input images while possessing the expression information.

\subsection{Attention Map Analysis}

\begin{figure}[th]
   \centering
   \includegraphics[width=0.5\textwidth]{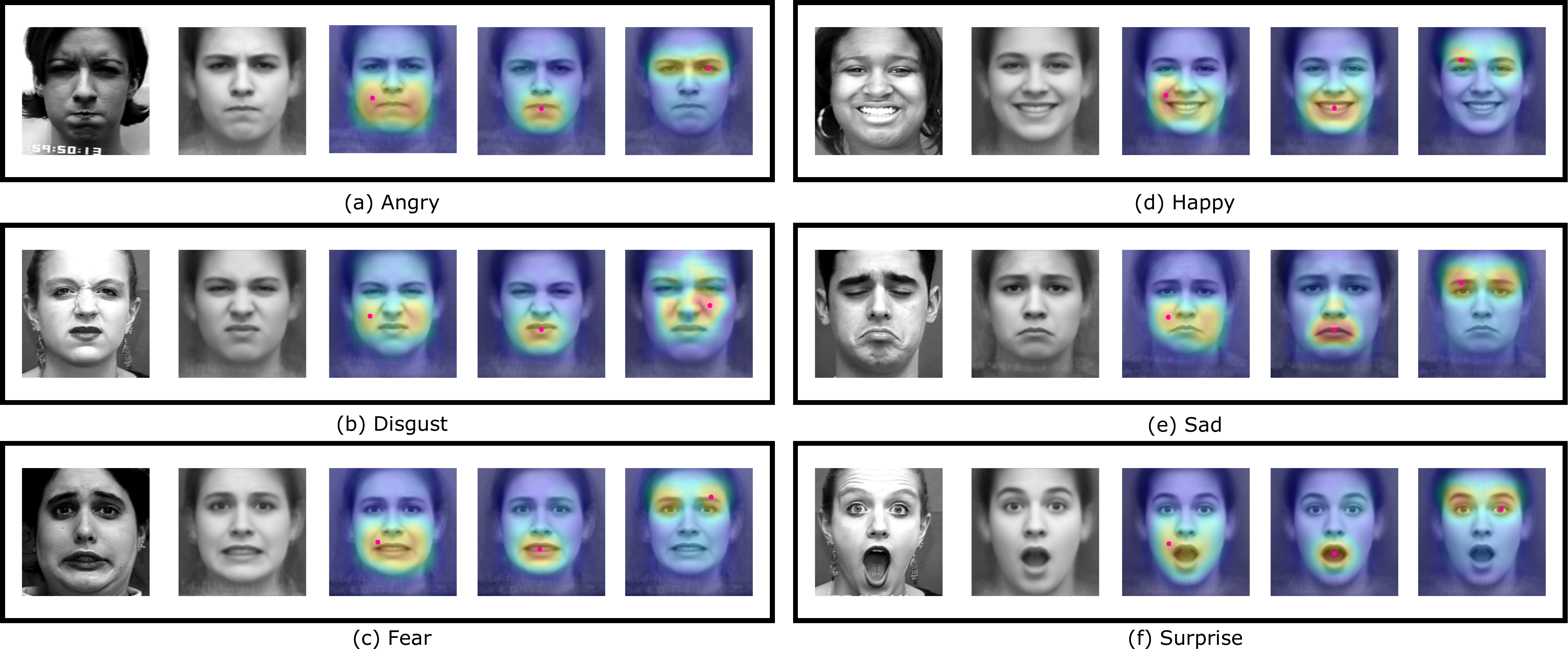}
   \caption{\small{Demonstration of long-range dependency learning across feature maps by visualizing the attention maps from the last U-Net layer self-attention module, the ones closest to the generated image. In each box, the first image is the input expressive image, the second image is the generated ``average" identity expressive image, and the final three images are learned attention maps for three query locations indicated by three dots. Best viewed in color.}}
   \label{fig:attention_result}
\end{figure}

To understand what self-attention modules learned after the $56 \times 56$ feature maps in Fig.~\ref{fig:U-Net}, we visualized the attention maps from the last U-Net layer self-attention module for different input images in Fig.~\ref{fig:attention_result}. We noticed that the self-attention modules learn to model the relationships among regions according to similarity of intensity, texture, and edge rather than just spatial adjacency. Furthermore, most of the attention maps have highlighted regions around the lip, nose, cheek, eyes, and eyebrows, consistent with the psychological studies~\cite{cohn1995computerized}. For example, in Fig.~\ref{fig:attention_result} (e), the three attention maps for three query locations focus on eye and eyebrow, lip, and cheek regions, respectively, and are highly related to a set of facial action units (AUs)~\cite{Ekman78}, which describe an expression. In Fig.~\ref{fig:attention_result} (d), regions highlighted for generating a happiness expression are located around the cheek,  AU6 (Cheek Raiser), and around the lip corners,  AU12 (Lip Corner Puller). In Fig.~\ref{fig:attention_result} (f), the regions related to a surprise expression~\cite{Lucey2010}, i.e., AU5 (Upper Lid Raiser) and AU27 (Mouth Stretch), are most highlighted. The self-attention module demonstrates similar observations for other expressions.

\section{CONCLUSION}

This work describes an end-to-end IF-GAN framework for identity-free FER. Differing from other subject-independent methods, IF-GAN removes identity-related information completely. Experimental results on four facial expression datasets show the proposed IF-GAN achieves state-of-the-art performance. Furthermore, results on RAF-DB show the IF-GAN can deal with variations in pose, occlusion, and illumination.

{\small
\bibliographystyle{IEEEbib}
\bibliography{../../../bibliography/abbrev_short,../../../bibliography/machine_learning/ty-literature_graphical_model,../../../bibliography/ty-literature_misc,../../../bibliography/ty-literature_self,../../../bibliography/emotion/ty-literature_AU_Exp_Emotion_rec,../../../bibliography/emotion/ty-literature_AU_rec,../../../bibliography/emotion/ty-literature_Exp,../../../bibliography/emotion/ty-literature_AU_rec1,../../../bibliography/emotion/ty-literature_emotion_rec_survey,../../../bibliography/ty-literature_audiovisual_ASR,../../../bibliography/ty-literature_facial_feature_detect_track,../../../bibliography/machine_learning/ty-literature_unsupervised_feature_learning,../../../bibliography/machine_learning/ty-literature_gan,../../../bibliography/ty-literature_database,../../../bibliography/machine_learning/ty-literature_machine_learning,../../../bibliography/machine_learning/ty-literature_deep_learning,../../../bibliography/machine_learning/ty-literature_metric_learning,../../../bibliography/ty-literature_statstical_models_alignment,../../../bibliography/object_classification/ty-literature_object_detection,../../../bibliography/ty-literature_psychology,../../../bibliography/machine_learning/ty-literature_loss,../../../bibliography/machine_learning/ty-literature_multi_task_learning,../../../bibliography/emotion/ty-literature_gan_exp,../../../bibliography/ty-literature_EmotiW,../../../bibliography/machine_learning/ty-literature_cnn_clustering,../../../bibliography/ty-literature_tools,../../../bibliography/ty-literature_attention,../../../bibliography/ty-literature_self}
}

\end{document}